\documentclass[10pt,journal,compsoc]{IEEEtran}

\ifCLASSOPTIONcompsoc
\usepackage{subcaption}
\usepackage{graphicx}
\usepackage{caption}
\usepackage{multirow}
\usepackage{adjustbox}
\usepackage{tabularx}
\usepackage{epstopdf}
\usepackage{amsmath} 
\usepackage{amssymb}
\usepackage{color}
\usepackage{algorithm}
\usepackage{algpseudocode}
\usepackage{scrextend}
\usepackage{todonotes}
\usepackage{lipsum}
\usepackage[nocompress]{cite}
\else
 \usepackage{cite}
\fi
\ifCLASSINFOpdf
\else
\fi

\def\bs{\boldsymbol}

\makeatletter
\newcommand{\removelatexerror}{\let\@latex@error\@gobble}
\makeatother

\begin{document}
\title{Dimension Estimation Using Autoencoders}

\author{Nitish~Bahadur and~Randy~Paffenroth
\IEEEcompsocitemizethanks{\IEEEcompsocthanksitem N. Bahadur is with the Department
of Data Science, Worcester Polytechnic Institute, Worcester,
MA, 01609.\protect\\
E-mail: nbahadur@wpi.edu
\IEEEcompsocthanksitem R. Paffenroth is the Associate Professor of Mathematical Sciences, Associate Professor of Computer Science, and Associate Professor of Data Science, Worcester Polytechnic Institute, Worcester,
MA, 01609.}
\thanks{Manuscript revised Sep 23, 2019.}}


\IEEEtitleabstractindextext{%
\begin{abstract}
	Dimension Estimation (DE) and Dimension Reduction (DR) are two closely related topics, but with quite different goals.  In DE, one attempts to estimate the intrinsic dimensionality or number of latent variables in a set of measurements of a random vector.  However, in DR, one attempts to project a random vector, either linearly or non-linearly, to a lower dimensional space that preserves the information contained in the original higher dimensional space.  Of course, these two ideas are quite closely linked since, for example, doing DR to a dimension smaller than suggested by DE will likely lead to information loss.  Accordingly, in this paper we will focus on a particular class of deep neural networks called autoencoders which are used extensively for DR but are less well studied for DE.  We show that several important questions arise when using autoencoders for DE, above and beyond those that arise for more classic DR/DE techniques such as Principal Component Analysis.  We address autoencoder architectural choices and regularization techniques that allow one to transform autoencoder latent layer representations into estimates of intrinsic dimension.          
\end{abstract}

}

\maketitle

\IEEEdisplaynontitleabstractindextext
\IEEEpeerreviewmaketitle

\ifCLASSOPTIONcompsoc
\IEEEraisesectionheading{\section{Introduction}\label{sec:introduction}}
\else
\section{Introduction}
\label{sec:introduction}
\fi


\IEEEPARstart{D}{imension} estimation (DE) is the process of determining the \emph{intrinsic dimensionality} of data (e.g., see Chapter 3 in \cite{nldrLeeVerlysen}).  Usually, real world datasets have large numbers of features, often significantly greater than the number of latent factors underlying the data generating process, and DE attempts to quantify the number of latent factors in a dataset.  Of course, latent factors are often discussed in relation a linear analysis.  However, such ideas can be generalized to a non-linear context, and such generalizations will be our focus here.  For example, in an image processing problem each pixel of an image can be thought of a feature that one measures about the image.  However, the measured pixels are not independent of each other since, for example, nearby pixels are likely to have similar colors.
Accordingly, it can be quite useful to estimate the underlying latent factors, such as pose and lighting, that effect many pixels simultaneously. 


The precise definition of a latent factor, and therefore the precise definition of the intrinsic dimensionality, can be quite challenging.  For example, Principal Component Analysis (PCA) (which we will discuss in detail in the sequel) defines latent factors in terms of linear projections and orthogonality.  At the other extreme, one can consider quite complicated scenarios involving fractal dimensions and space filling curves \cite{nldrLeeVerlysen}.  In this text we take a middle of the road approach, and focus on non-linear, but smooth, manifolds.  
Such an approach is quite popular, and such DR and DE techniques are used in diverse domains such as engineering, astronomy\cite{hoyle2015anomaly}\cite{baron2016weirdest}, biology\cite{dai2006dimension}\cite{parsons2017dimension}, remote sensing\cite{zhao2016spectral}\cite{dalponte2008fusion}, economics\cite{nassirtoussi2015text}\cite{rathnayaka2013econometric}, social media\cite{nori2012multinomial}\cite{lee2017big}, and finance\cite{wang2006effects}\cite{albanese2002dimension}; and this class of techniques has a large extant literature (see, e.g., \cite{van2009dimensionality, nldrLeeVerlysen} and references therein).



{\textit{Dimension Reduction} (DR) is the process of reducing a high dimension dataset with $N$ features into a dataset with $p$ features, where $p \ll N$. 
Perhaps the most classic method in this domain is PCA\cite{jolliffe2002principal}, and a detailed study of PCA will illuminate many of the issues that arise when performing DR and DE with autoencoders.  Unfortunately, most DR methods need an estimate of the number of latent variables as a user-defined input to the process of dimensionality reduction.  Estimates of the number of latent variables in a particular dataset can come from a variety of principles (e.g., see Chapter 3 in \cite{nldrLeeVerlysen}) and looking at the example of PCA will provide illumination for our work.

In particular, as we will foreshadow here and detail in the sequel, PCA is commonly implemented using the singular-value decomposition(SVD) \cite{golub1970singular} where a data matrix $\bs{X}\in\mathbb{R}^{m \times n}$ is factored into the form ${\bs{U\Sigma V^{T}}}$, where $\bs{U}\in\mathbb{R}^{m \times m}$ is an unitary matrix, $\bs{\Sigma}\in\mathbb{R}^{m \times n}$ is rectangular diagonal matrix with non-negative real numbers on the diagonal, and $\bs{V}\in\mathbb{R}^{n \times n}$ is also an unitary matrix. The diagonal entries ${\bs{\sigma _{i}}}$ of ${\bs{\Sigma}}$ are known as the singular values of ${\bs{X}}$.  Dimension reduction to dimension $k$ for $\bs{X}$ can be performed by removing the $n-k$ columns of ${\bs{\Sigma}}$ with smallest ${\bs{\sigma _{i}}}$, giving rise to a ${\bs{\hat{\Sigma}}}\in\mathbb{R}^{m \times k}$, and then computing $\bs{U} \bs{\hat{\Sigma}}$.  
Of course, in the presence of finite samples and noise in data the choice of an appropriate $k$ is non-trivial.  In particular, any ${\bs{\sigma _{i}}}$ with the property that
${\bs{\sigma _{i}}}=0$ can be removed without changing the properties of the data (e.g., the Euclidean distances between the points).  However, in real-world data it is rarely the case that 
${\bs{\sigma _{i}}}=0$ for any ${\bs{\sigma _{i}}}$, and dimension reduction using PCA will change the properties of the data in $\bs{X}$.

As a running example in this paper, consider images of hand-written digits from the MNIST\cite{deng2012mnist} dataset.  For example, using PCA and a user-defined estimate of the number of latent variable $k = 50,100,200,400, 784$ we reduce the number of features of digit $0$ and reconstruct digit $0$ using the formula $\bs{\hat{X}} = \bs{U} \bs{\hat{\Sigma}} \bs{\hat{V}^T}$ for
$\bs{\hat{V}}\in\mathbb{R}^{n \times k}$, to see the effect of various levels of dimension reduction. Visually, even when we reduce the number of dimension of digit 0 from 784 to 50, the reconstructed digit $0$ is visually (figure~\ref{fig:mnist_k_dim}) similar to the original digit 0.  Note, already a quite important issue has arisen.  How is one to chose $k$? To assist the latent dimension estimation process a scree plot (figure~\ref{fig:mnist_digit0_pca}) of the PCA digit $0$, which plots the sizes of the singular values ${\bs{\sigma _{i}}}$, can be used. Moreover, one may expect that reconstruction error will remain low if $k$ is greater than the intrinsic dimensionality of the linear PCA embedding. On the contrary, if $k$ goes below the intrinsic dimensionality, the dimensionality reduction may cause a sudden increase in reconstruction error. This is illustrated in digit $0$ using a reconstruction error plot(figure~\ref{fig:mnist_digit0_pca_recon_err}) as $k$ decreases.

\begin{figure}[htp]
    \centering
    \includegraphics[width=3.2in]{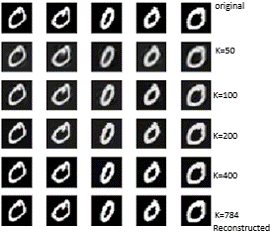}        
    \caption{Each MNIST digit $0$ is ${28 \times 28}$ pixel image that gives us $784$ features.  We arbitrarily choose a starting dimensionality $k=50$ and reduce the number of features from $784$ down to $50$ and reconstruct and plot digit $0$. This process is repeated by increasing $k$ till $k=784$, the maximum number of features.} 
    \label{fig:mnist_k_dim}
\end{figure}

\begin{figure}[htp]
    \centering
    \includegraphics[width=3.2in]{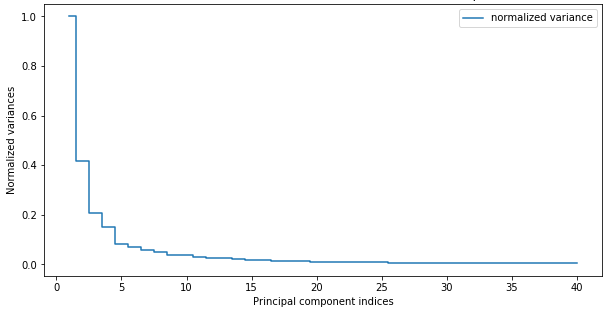}        
    \caption{Scree plot of normalized variance of digit $0$ shows that as the number of principal components increase the normalized variance quickly drops before tapering out at $30$, a number much less than original $784$ components.  Hence, a possible approximation of the intrinsic dimension might be $30$.} 
    \label{fig:mnist_digit0_pca}
\end{figure}

\begin{figure}[htp]
    \centering
    \includegraphics[width=3.2in]{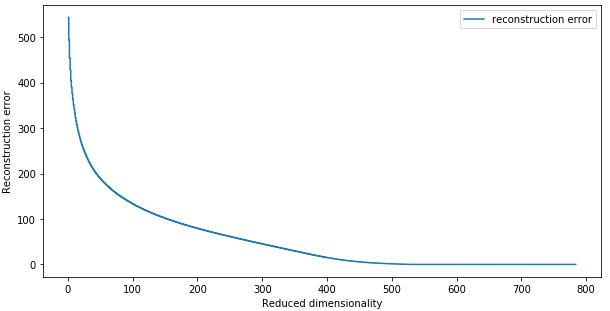}        
    \caption{Estimation of intrinsic dimensionality $k$ of digit 0 as the number of dimension increases.  When $k$ is reduced below 50 there is sudden increase in reconstruction error.} 
    \label{fig:mnist_digit0_pca_recon_err}
\end{figure}


%

However, we are most interested in moving beyond the linear DR provided by PCA and estimating non-linear dimensionality of datasets.  Therefore, we study how autoencoders\cite{rumelhart1985learning}\cite{vincent2008extracting}, a classic deep neural network used for DR, can also be used for DE.  



An autoencoder (AE) is a type of artificial neural network used to learn an approximation to the identity function, so that the output {\textbf{\^{X}}}  is similar to n-dimensional input $\bs{X}$. Because the hidden layer acts a bottleneck, when the autoencoder compresses the input to a latent space representation and then reconstructs the output from the latent space representation the hidden units encode significant features in input data $\bs{X}$.

\begin{figure}[htp]
        	\centering
        	\includegraphics[width=3.2in]{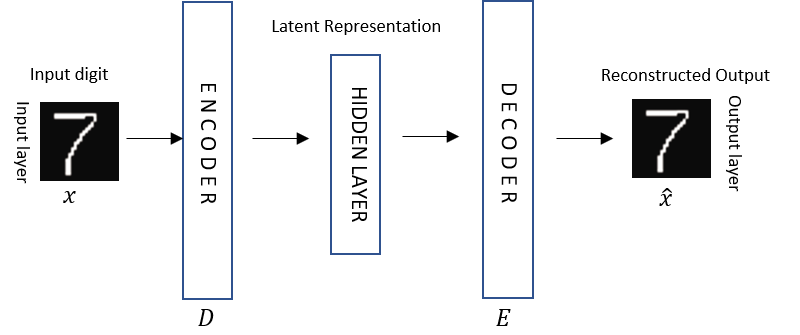}        
        	\caption{A schematic of an autoencoder with input layer, encoding layer, hidden layer, decoding layer and output layer.}
	\label{fig:autoencoder}
\end{figure}

%


Just like $k$, which defines the number of principal components in PCA, $n$ (figure~\ref{fig:mnist_ae_diff_hidden}), which defines the number of nodes in the innermost hidden layer, needs to be estimated before building the autoencoder.  Again, how do we estimate the appropriate number of nodes in the innermost hidden layer?  PCA and autoencoders are closely related\cite{plaut2018principal}.  An autoencoder with an identity activation function, and therefore linear layers, and a squared loss function will compute the same subspace as PCA\cite{plaut2018principal}.  However, the parameterization of the subspace can be quite different.  For example, the SVD guarantees that the basis is orthogonal and the singular values are ordered.  \emph{Running gradient descents, a common optimization routine for neural networks, on an autoencoder makes no such guarantees.  Restoring such guarantees for an autoencoder is the focus of our work.}

\begin{figure}[htp]
    \centering
    \includegraphics[width=3.2in]{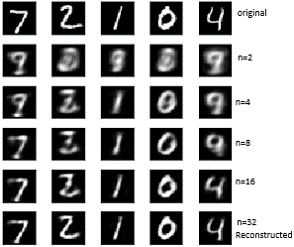}        
    \caption{Similar to figure~\ref{fig:mnist_k_dim} this figure shows the reconstruction performance as the number of nodes($n$) in the autoencoder hidden layer increase from 2 to 32 the reconstructed image visually resembles more and more like the input.} 
    \label{fig:mnist_ae_diff_hidden}
\end{figure}


AEs are designed for dimension reduction, but we are interested in using AE for estimating \emph{intrinsic dimension}.  Unlike PCA, AEs do not natively support dimension estimation.  There are no ordered singular values equivalent in the hidden layer that can be used to estimate dimension.  For example, if we were to run the previous input (figure~\ref{fig:mnist_k_dim}) through an autoencoder with 32 nodes in innermost hidden layer (figure~\ref{fig:autoencoder}) and scatter plot the hidden layer for digit $0$ we see a set of values with no clear pattern that can be used to estimate the dimension of digit $0$, as illustrated in figure~\ref{fig:z_layer_orientation_64_0}.  To transform the AE hidden layer into singular value proxies and subsequently use the proxies to estimate dimension is our research.

\begin{figure}[htp]
    \centering
    \includegraphics[width=3.2in]{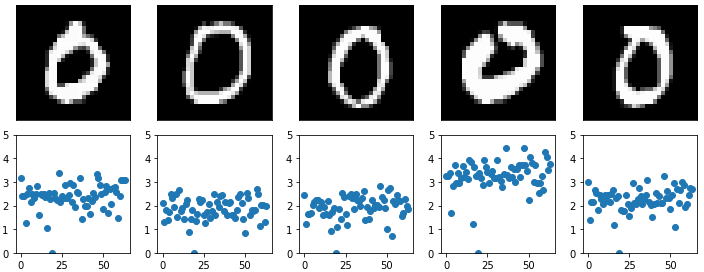}        
    \caption{Here we show a selection of images of $0$ and the $64$ dimension innermost hidden layers that correspond to each image.  Note that for each image the innermost hidden layer has all $64$ entries non-zero.   Does that mean the intrinsic dimension is therefore $64$?  Not necessarily, since there is not penalty on the innermost hidden layer to force entries to be $0$ in a classic AE.}
    \label{fig:z_layer_orientation_64_0}
\end{figure}


\subsection{CONTRIBUTIONS}\label{sec:contributions}

The main contributions of the paper can be summarized as follows:  

\begin{enumerate}
  \item We designed an algorithm to estimate the non-linear intrinsic dimension using AEs.  The values of the innermost hidden layer are not appropriate for DE.  However, we develop appropriate singular value proxies (SVP) that allow the dimension to effectively estimated.
  \item We applied dimension estimation algorithm to time series dataset such as stock prices of S\&P 500 index constituents as well as image data such as the MNIST digits dataset.
  \item We compared and contrasted how dimension estimation differs between linear PCA and a non-linear autoencoder.
\end{enumerate}

\subsection{BACKGROUND}\label{sec:background}
While quantifying intrinsic dimension using linear technique such as PCA is standard, estimating the dimension of real-world time series datasets using autoencoders is more challenging.  For example, Wang, Yao, and Zhao\cite{wang2016auto} used an autoencoder with varying number of nodes in hidden layer MNIST and Olivetti\cite{samaria1994parameterisation} dataset to determine intrinsic dimension. The authors find that as the number of nodes in hidden layer increased, the classification accuracy increased before plateauing.  Our approach is fundamentally different.  First, we do not change the network architecture (number of hidden nodes), since we found that using a binary search method to estimate the number of nodes required in hidden layer was quite inefficient. Second, in our approach each class of digits in the MNIST dataset has a different dimension and we have no reason to conjecture that the number of class labels (10 for MNIST) should be the same as intrinsic dimension of each digit.

Recently (2018), Li, Farkhoor, Liu, and Yosinski \cite{li2018measuring} used a 2 layer fully connected (FC) and convolutional neural network (CNN) classifier to measure the intrinsic dimension of a neural network itself. For MNIST at 90\% classification accuracy the network's intrinsic dimension in case of FC was 750 and CNN was 290. Although the research is quite impressive, our goal is not determining intrinsic dimension of neural networks, but rather estimate intrinsic dimension of the manifold on which the MNIST digits lay.

A widely studied problem in finance is stock portfolio diversification and such problems are closely connected to DE \cite{alexeev2012equity}~\cite{statman1987many}~\cite{tang2004efficient}.  However, as far as we know, there is scant research studying the non-linear intrinsic dimension of financial markets.   This motivates us to estimate the dimension of financial markets as one of our examples herein.

This paper is structured through six sections. In section~\ref{sec:introduction}, we provide the problem definition and a brief introduction of relevant existing literature. We provide an overview of dimension reduction, building blocks to estimate dimension, in section~\ref{sec:pca_mds_iso_ae}.  Section~\ref{sec:ae_svp} elaborates architectural choices we make in designing autoencoder for dimension estimation that allow for creating SVPs for autoencoders.  Once we have SVPS then section~\ref{sec:dd_algo} details an algorithm to quantify the SVPS into dimensionality estimates. Section~\ref{sec:experiment} presents our experiments with MNIST and S\&P 500 datasets.  Additionally, section \ref{sec:experiment} compares and contrasts the differences between PCA and autoencoder.  Finally,  section~\ref{sec:conclusion} provides a summary and pointers to future work.

\section{Dimension Reduction}\label{sec:pca_mds_iso_ae}
In this section, we provide an overview of linear dimension reduction technique such as PCA and nonlinear dimension reduction technique such as Isomap \cite{tenenbaum2000global} and autoencoders. 

\subsection{Linear Dimension Reduction Techniques}\label{sec:linear_dr}
Linear techniques such as Principal Component Analysis (PCA)\cite{hotelling1933analysis} have seen wide use.  The key idea of PCA is to construct low-dimensional sub-spaces that preserve as much of the variance in the data as possible and thereby preserve the data's correlation structure.

\subsubsection{Principal Component Analysis}\label{sec:pca}
As mentioned previously, PCA is a linear dimension reduction technique widely used in machine learning.  Using correlations between features, PCA finds the direction of maximum variance in high dimensional data and projects data onto a new subspace of fewer dimension. \cite{hotelling1933analysis} 
As mentioned previously, we assume that we have a data matrix $\bs{X} \in \mathbb{R}^{m\times n}$ where each of the $m$ rows is thought of as a point in $m$ dimensional space.  

The \textit{singular value decomposition} (SVD) (figure~\ref{fig:svd}) of a $\bs{X}$ is 
\begin{equation}
\bs{X}=\bs{U \Sigma V^T}
\end{equation}

\noindent where,
\begin{itemize}
  \item $\bs{V}\in \mathbb{R}^{m\times m}$ is an orthonormal (or unitary) matrix such that $\bs{V^T V} =\bs{I}_{m\times m}$ (an $m \times m$ identity matrix).
  \item $\bs{\Sigma}$ is a pseudodiagonal matrix with the same size as $\bs{X}$; the $m$ entries $\bs{\sigma}_m$ on the diagonal are called the singular values of $\bs{X}$.
  \item $\bs{U}\in \mathbb{R}^{n\times n}$ is an orthonormal (or unitary) matrix such that $\bs{U^T U} =\bs{I}_{n\times n}$ (an $n \times n$ identity matrix).
\end{itemize}

\begin{figure}[htp]
  \centering
  \includegraphics[width=3.2in]{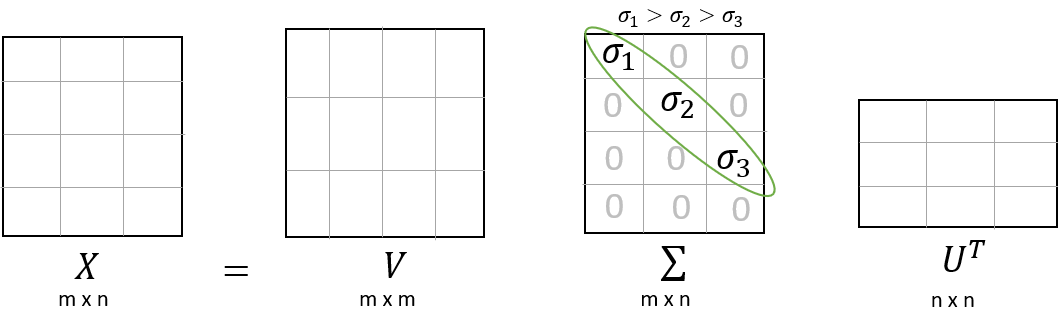}        
  \caption{Singular value decomposition of $\bs{X}=\bs{V}\bs{\Sigma}\bs{U}^T$.  The singular values ${\bs{\sigma _{i}}}$ are ordered from largest to smallest.}
  \label{fig:svd}
\end{figure}

The linear low-dimensional project of $X$ to $k$ dimension space is
$\bs{U} \bs{\hat{\Sigma}}$ where $\bs{\hat{\Sigma}}$ has its rightmost $n-k$ columns removed,
corresponding to the $n-k$ smallest singular values ${\bs{\sigma _{i}}}$.  Note, PCA provides two important features that we will use in the sequel.  Namely, the columns of $\bs{U}$ are orthonormal and the singular values ${\bs{\sigma _{i}}}$ are ordered from largest to smallest.

\subsection{Nonlinear Dimension Reduction Technique}\label{sec:nonlinear_dr}
Real world data does not always lay on a low-dimensional linear manifold, but can instead be most naturally represented by low-dimensional non-linear manifold.  There is a large extent literature on such methods, which is beyond the scope of the current text.  However, the interested reader can look to \cite{nldrLeeVerlysen}, and references therein, for details on such methods.  However, we compare our AE based technique with an important method in this domain called \emph{Isomap} \cite{tenenbaum2000global}.  In particular, we will compare the non-linear DE capabilities of Isomap with our proposed autoencoder based algorithm, so in this section we provide a brief overview of Isomap and more in-depth derivation of autoencoders.

\subsubsection{Isomap}
Isomap~\cite{nldrLeeVerlysen} \cite{yang2002extended}\cite{balasubramanian2002isomap} maps data points in high-dimensional nonlinear manifold to a lower dimensional set of coordinates.  It successfully addresses important limitations in MDS\cite{kruskal1964multidimensional} by using geodesic distance - distance along the curve- between two given points - instead of Euclidean distance - which is the straight-line distance between corresponding nodes.  The algorithm works by creating a neighborhood and neighbors are converted into a graph structure.  In particular, the steps of the algorithm are as follows.

\begin{enumerate}
  \item For each row in $\bs{X}\in\mathbb{R}^{m \times n}$ considered as a point in $\mathbb{R}^m$, choose $k$ nearest points as neighbors \cite{friedman1977algorithm}
  \item Consider all the point in $\bs{X}$ as nodes and if any two nodes are chosen to be neighbors in 1), calculate Euclidean distance between them $\bs{D}=[d_{ij}^2]_{n\times n}$; where $d_{ij} = \|\bs{x}_i - \bs{x}_j\|_2$ and $n$ is the order of the high-dimensional space. This step converts the dataset into a graph.
  \item For each pair of nodes in the graph, find the points $\mathcal{G}=\big\{\bs{x}_i\vert i=1 \dots, k\big\}$ in the shortest path using Floyd's algorithm \cite{floyd1962algorithm} and assign it to $\bs{D}_{ij}$ for that non-neighnoring pair of nodes.
  \item Convert the matrix of distances $\bs{D}$ into a Gram matrix $\bs{S}$ by double centering using $S_{ij}=-\frac{1}{2}\big[d^2_{ij}-\mu_i(d^2) -\mu_j(d^2)+\mu_{ij}(d^2)\big]$, where $\mu_i$ is the average of the $i$-th row, $\mu_j$ is the average of the $j$-th column, and $\mu_{ij}$ is the average of all of the entries in the matrix.  Note, $S$ is symmetric semi-positive definite if $D$ is a Euclidean Distance Matrix \cite{nldrLeeVerlysen}. 
  \item Compute the SVD of $S$ using $\bs{S}=\bs{U}\bs{\Sigma}\bs{V}^T$.
  \item Finally, estimate $k$ dimensional latent variables as $\hat{\bs{X}}=\bs{I}_{k\times n}\bs{\Sigma}^{1/2}\bs{V}^T$ where $\bs{I}_{k\times n}$ is a $k\times n$ matrix with $1$ on its diagonal and $0$ elsewhere.
\end{enumerate}

\subsubsection{Autoencoder}\label{sec:ae}
AEs are, in many ways, quite similar to PCA.  In particular, at their simplest, an
autoencoder with one layer hidden takes input data $\bs{x}_i \in \mathbb{R}^{n\times 1}$ as transforms that vector into a new $\bs{y}_i \in \mathbb{R}^{k\times 1}$, often called the hidden layer, according to the mapping 

\begin{equation}\label{eqn:y_i}
\bs{y}_i  = \sigma(W_1\bs{x_i} + \bs{b_1})
\end{equation}

\noindent where $W_1$ is the weight matrix of the first layer, and $k < n$.  The function $\sigma$ is typically a non-linear activation function such as a sigmoid or a rectified linear unit (ReLU) \cite{nair2010rectified}.  Another layer then maps  $\bs{y_i}$ to $\hat{\bs{x_i}} \in \mathbb{R}^{n\times 1}$ according to 

\begin{equation}\label{eqn:x_hat}
\hat{\bs{x_i}} = \sigma(W_2\bs{y_i} + \bs{b_2}) = \sigma(W_2 \sigma(W_1\bs{x_i} + \bs{b_1}) +\bs{b_2})
\end{equation}

\noindent where $W_2 \in \mathbb{R}^{n\times k}$ and $\bs{b}_2 \in \mathbb{R}^{n\times 1}$ are the weight matrix and bias vector of second layer.  Deep networks with many layers can be defined in an analogous fashion.  The parameters $W_1$, $\bs{b_1}$, $W_2$, $\bs{b_2}$ are found by minimizing some cost function(eqn~\ref{eqn:ae_cost_fn}) that quantifies the difference between the output $\hat{\bs{x_i}}$ and input as in

\begin{equation}
J(W,b;x) = \frac{1}{2}\|x - \hat{x}\|_2^2,
\label{eqn:ae_cost_fn}
\end{equation}

\noindent with $W=\{W_1,W_2\}$ and $b=\{b_1,b_2\}$, leading to the optimization problem

\begin{equation}
  \min_{W,b} J(W,b;x).
  \label{eqn:ae_cost_fn_opt}
\end{equation}
  
At this point, we can begin to understand the differences between PCA and autoencoder.  In particular, $\bs{y_i}$ is a projection of $\bs{x_i}$ into a $k$ dimensional space, just as
$\bs{U} \bs{\hat{\Sigma}}$ is a projection of the rows of a data matrix $X$ into a $k$ dimensional space.  However, that is where the analogy ends.  While the basis in $\bs{U}$ is explicitly orthonormal and the diagonal entries in $\bs{\hat{\Sigma}}$ are ordered by size, the structure of the autoencoder is hidden inside the nonlinear function $\sigma(W_1\bs{x_i} + \bs{b_1})$.  \emph{So, as opposed to the singluar values in $\bs{\Sigma}$, the size of the entries in $\bs{y_i}$ do not provide information about the intrinsic dimensionality of the data set.}

Fortunately, the ideas in \cite{ng2011sparse} provide a path forward.  When a neural network is constrained by $k$ being small, then the network is forced to learn a compressed representation of input.  However, when $k$ is not known beforehand, which is the entire point of DE, one can penalize the hidden layer entries $\bs{y_i}$ so that they are encouraged to be small, or even $0$.  As shown in figure~\ref{fig:z_layer_orientation_64_0_l1}, to achieve sparsity hidden layer is penalized with $\|.\|_1$.  $\|.\|_1$ forces small hidden layer entries to zero and prevents over estimation of dimension.

\begin{figure}[htp]
    \centering
    \includegraphics[width=3.2in]{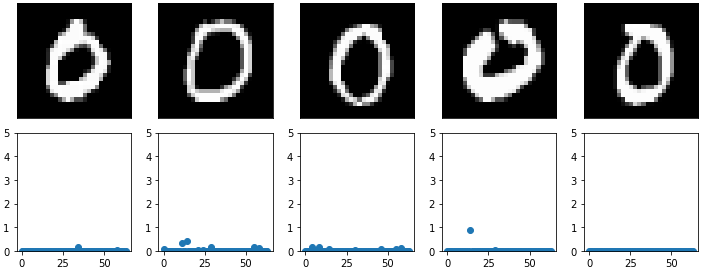}        
    \caption{Here we show a selection of images of $0$ and the $64$ dimension innermost hidden layers with $\|.\|_1$ that correspond to each image.  $\|.\|_1$ penalty on hidden layer entries makes the entries closer to 0.}
    \label{fig:z_layer_orientation_64_0_l1}
\end{figure}

In that way, the number of large entries in $\bs{y_i}$ can be interpreted as a measure of the intrinsic dimension.  

Additionally, to achieve consistency in DE and make the DE process independent of range of numeric values we use $\|.\|_2$, which normalizes the values (figure~\ref{fig:z_layer_orientation_64_0_l2}).

\begin{figure}[htp]
    \centering
    \includegraphics[width=3.2in]{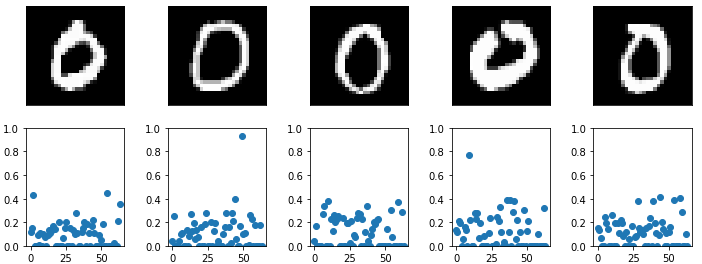}        
    \caption{Here we show a selection of images of $0$ and the $64$ dimension innermost hidden layers with $\|.\|_2$ that correspond to each image.  $\|.\|_2$ normalizes hidden layer entries.}
    \label{fig:z_layer_orientation_64_0_l2}
\end{figure} 

DE requires both sparsity and consistency, which is shown in figure~\ref{fig:unsorted_digit_0} and figure~\ref{fig:row_sorted_digit_0}.

Accordingly, we follow the ideas in 
\cite{ng2011sparse} and start by normalizing the hidden layer as in eqn~\ref{eqn:y_i_normalize} 

\begin{equation}\label{eqn:y_i_normalize}
y_{norm}^i = \frac{y_{i}}{\sqrt{\sum_{i=1}^k {y_{i}}^2}}
\end{equation}

\begin{equation}\label{eqn:recon_err}
J_{sparse}(W,b;x) = J(W,b;x) + \lambda{\sum_{i=1}^k \| y_{norm}^i\|_1}
\end{equation}

\noindent to force some entries of $\bs{y_i}$ to be small or even $0$.  The number of large entries in $\bs{y_i}$, when appropriately interpreted as described in the sequel, the provide an estimate of the dimensionality of $X$. 

\subsubsection{Autoencoder Model}\label{sec:ae}
To estimate dimension of MNISTdigits the AE model uses 5 layer autoencoder.  Table~\ref{tbl:mnist_ae_arch} defines the layers, activation function and regularizers. The input and output layer has 784 nodes; the output layer uses sigmoid \cite{cybenko1989approximation} activation function. To estimate dimension of financial market, using S\&P 500 index constituents, we again use 5 layer autoencoder as illustrated in table~\ref{tbl:sp500_ae_arch}. The input and output layer has 550\footnote{S\&P 500 index constituents change due to merger,  acquisitions or new public listed companies being added to index.} nodes and uses tanh\cite{glorot2011deep} activation function. 

\begin{table}[ht]
    \begin{center}
        \begin{tabular}{cccc}
            \hline
	Layer Type & MNIST & Regularizer & Activation Function \\\hline
      input layer & 784 & - & - \\\hline
      encoder layer 1 & 256 & - & relu \\\hline        
      encoder layer 2 & 128 & - & - \\\hline        
      hidden layer & 64 & $l1l2$ (eqn~\ref{eqn:recon_err}) & - \\\hline        
      decoder layer 1 & 128 & - & - \\\hline        
      decoder layer 2 & 256 & - & relu \\\hline        
      output layer & 784 & - & sigmoid \\\hline        
    \end{tabular}      
    \end{center}
    \caption{MNIST autoencoder architecture.}
    \label{tbl:mnist_ae_arch}
\end{table}

\begin{table}[ht]
    \begin{center}
        \begin{tabular}{cccc}
            \hline
	Layer Type & S\&P 500 & Regularizer & Activation Function \\\hline
      input layer & 550 & - & - \\\hline
      encoder layer 1 & 256 & - & relu \\\hline        
      encoder layer 2 & 128 & - & - \\\hline        
      hidden layer & 64 & $l1l2$ (eqn~\ref{eqn:recon_err}) & - \\\hline        
      decoder layer 1 & 128 & - & - \\\hline        
      decoder layer 2 & 256 & - & relu \\\hline        
      output layer & 500 & - & tanh \\\hline        
    \end{tabular}      
    \end{center}
    \caption{S\&P 500 autoencoder architecture.}
    \label{tbl:sp500_ae_arch}
\end{table}

\section{Autoencoder - Singular Value Proxies}\label{sec:ae_svp}
SVD decomposes input matrix $\bs{X}$ such that $\bs{X}=\bs{V}\bs{\Sigma}\bs{U}^T$, where $\bs{V}$, $\bs{U}$ are unitary matrices and  $\bs{\Sigma}$ contains the singular values along the diagonal , where the largest value in upper left and smallest value in lower right.  Since we are primarily interested in DE using AE innermost hidden layer, which does not natively provide singular value equivalent, we need to create \emph{singular value proxies} to use innermost hidden layer for DE.  


\subsection{Number of rows}\label{sec:ae_width}
The choice of width, \emph{number of rows}, is driven by a few mathematical and practical factors.


In a time series data set, such as stock prices, choice of width is driven by temporal relevance and availability of data.  Estimating \emph{width} is tricky.  For MNIST digits we found that DE algorithm converged when we increased \emph{width}.  DE convergence is illustrated in figure~\ref{fig:digit_0_width} and figure~\ref{fig:digit_1_width}.  For brevity we only present results for digit 0 and 1 where width is monotonically increased from 2 to 90.  For each width we ran the experiment 50 times and then estimated average dimension.
\begin{figure}[htp]
        	\centering
        	\includegraphics[width=3.2in]{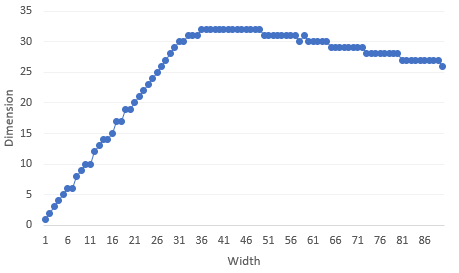}        
        	\caption{DE of digit 0 increases with width and then plateaus.}
	\label{fig:digit_0_width}
\end{figure}
\begin{figure}[htp]
        	\centering
        	\includegraphics[width=3.2in]{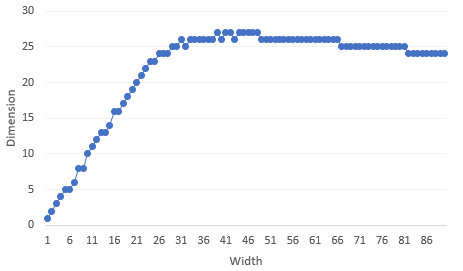}        
        	\caption{Similar to DE of digit $0$, the DE of digit $1$ also increases with width and plateaus.}
	\label{fig:digit_1_width}
\end{figure}

\subsection{L2/L1 Hidden Layer Regularizer}\label{sec:ae_norm}
Our goal is to use AE innermost hidden layer entries to estimate dimension.  While AE faithfully reproduces the input we have no control on the transformations in innermost hidden layer.  To make DE consistent and small, we use $\|.\|_{2,1}$ penalty on hidden layer entries.


\subsection{Activation Function}\label{sec:ae_actv_fn}
When autoencoder is used for DE we should avoid using activation function just before the innermost hidden layer. Relative magnitude between scalar value proxies change when activation function output is used for DE.  For example, the ratio of $\sigma_1$ and $\sigma_2$ in illustrated example (figure~\ref{fig:act_fn_eff}) is 10. Contrastingly, when sigmoid activation function is used the ratio changes to 1 and consequently affects our singular value proxies and DE.

\begin{figure}[htp]
        	\centering
        	\includegraphics[width=3.2in]{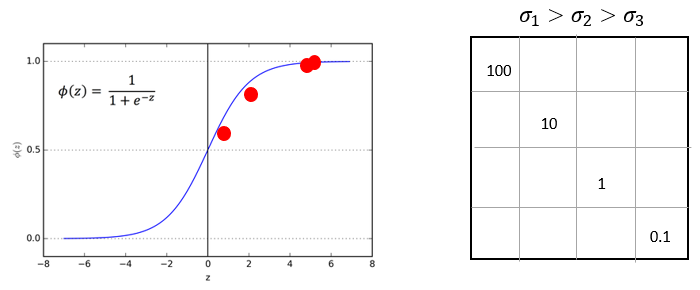}        
        	\caption{The relative scale of the scalar values change when we use an activation function.  }
	\label{fig:act_fn_eff}
\end{figure}

The effect of activation function on digit $0$ on MNIST test dataset is illustrated in figure~\ref{fig:noactfn_and_actfn_z_layer_64}.  The  innermost hidden layer absolute values range between 0 and 1.  However, when no activation function is used innermost hidden layer absolute values range between 0 and 6.

\begin{figure}[htp]
        	\centering
        	\includegraphics[width=3.2in]{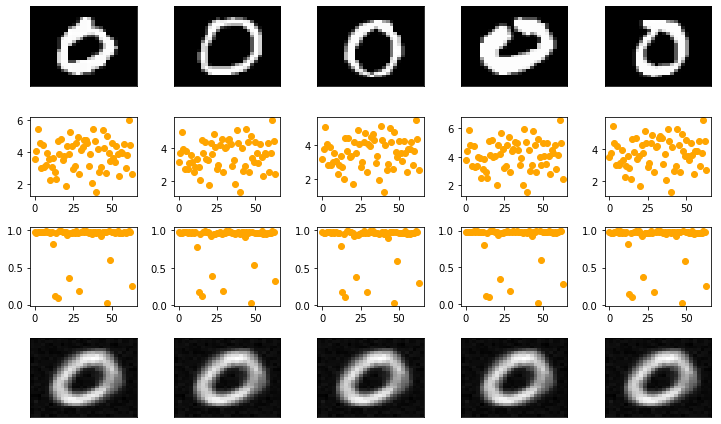}        
        	\caption{In the innermost hidden layer, whether activation function is used or not, AE output is invariant.  AE is designed to learn the input and produce an output similar to input.  As illustrated in row $2$ and $3$, the innermost hidden layer values are very different with and without an activation function (sigmoid in this case).  For DE we use the innermost hidden layer values from AE to estimate dimension and hence we do not use activation function in the innermost hidden layer.}
	\label{fig:noactfn_and_actfn_z_layer_64}
\end{figure}

%

Activation function output should not be used for DE. However, this design decision does neither affects the choice of activation function for other layers nor how many layers can a AE have.


\subsection{Sparsity Parameter}\label{sec:ae_sprs_param}
Deep autoencoder have thousands of parameters to train.  AE is data hungry and computationally intensive.  We used an iterative approach to determine $\bs{\lambda}$ the sparsity parameter in our experiment, as shown in figure~\ref{fig:bear_08112011_raw_zlayer_diff_lambda} and figure~\ref{fig:bull_10052015_raw_zlayer_diff_lambda}.  For example, if you were to use S\&P 500 stocks over last 10 year as input we will have a mere 2520 end of day prices (252 trading days x 10 year).  

\subsection{Latent Space Sorting}\label{sec:ae_lss}
To create singular value proxies we use \emph{Transforming Latent Representation to Singular Value Proxies} (algorithm~\ref{fig:ae_row_sort}).  The algorithm transforms entries of latent layer into singular value proxies which is used for estimating dimension of dataset, as illustrated in figure~\ref{fig:rs_col_avg}.

\begin{figure}[htp]
        	\centering
        	\includegraphics[width=3.2in]{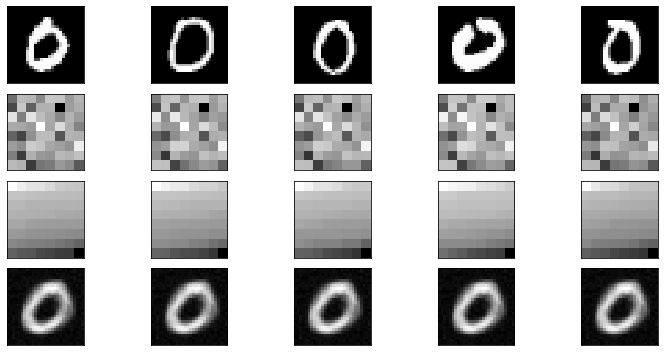}        
        	\caption{Without any sorting, innermost hidden layer $64$ values for digit $0$ from MNIST test dataset for first $5$ digit $0$ in shown in a $8 \times 8$ image.  Innermost hidden layer $64$ values for digit $0$ from MNIST test dataset for first $5$ digit $0$ in shown in a $8 \times 8$ image after sorting row wise, where for the innermost hidden layer values are arragned from largest to smallest.}
	\label{fig:latent_row_unsorted_and_sorted}
\end{figure}

%

\begin{figure}[htp]
        	\centering
        	\includegraphics[width=3.2in]{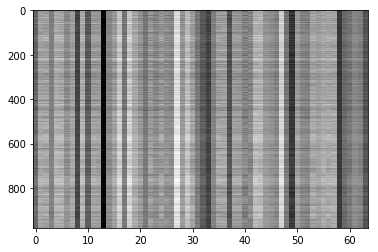}        
        	\caption{Without any transformation, the innermost hidden layer, which has $64$ nodes, for all the $980$ digit $0$ of MNIST test dataset is shown here.}
	\label{fig:unsorted_digit_0}
\end{figure}

\begin{figure}[htp]
        	\centering
        	\includegraphics[width=3.2in]{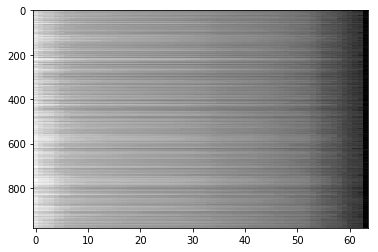}        
        	\caption{After sorting $64$ innermost hidden layer values for each $980$ digit $0$ we get an image where the larger values are on the left and smaller values are on the right.}
	\label{fig:row_sorted_digit_0}
\end{figure}

\begin{figure}[htp]
        	\centering
        	\includegraphics[width=3.2in]{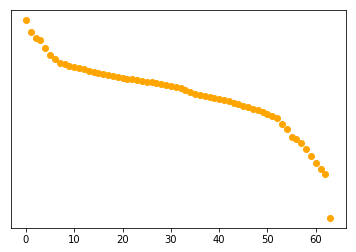}        
        	\caption{Take the column wise mean after sorting rows.}
	\label{fig:z_mean}
\end{figure}

\begin{algorithm}[!htp]
\caption{ \textit{Transforming Latent Representation to Singular Value Proxies.
\\ Inputs: The hidden layer $\bs{Z}$.  The matrix $Z$ is $mxn$.
\\Output: $\sigma_i$, a set of positive values that act as singular value proxies and is used to estimate the dimension of dataset . }}
\begin{algorithmic}[1]
    \State We need positive values to estimate dimension.  So create $\bs{M = abs(Z)}$
    \State Sort each row of $M_{sorted} = row\_sort(M)$ independently such that the largest value of each row is on the left and smallest value of each row is on right
    \State Take average of each column of $M_{sorted}$.  
    \State Calculate sum of each columns as $\bs{{z}_{mn}}=\bs{\sum_{i=1}^m} {z_{in}}$
    \State Singular value proxies is the average of column values calculated as $\bs{z_n}=\bs{\frac{z_{mn}}{m}}$
\newline
\end{algorithmic}\label{fig:ae_row_sort}
\end{algorithm}

\begin{figure}[htp]
        	\centering
        	\includegraphics[width=3.2in]{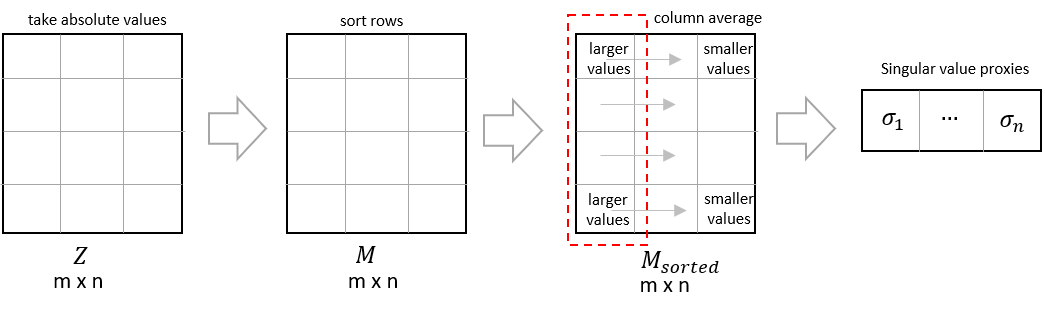}        
        	\caption{Steps needed to convert a latent representation into singular value proxies that can be used to estimate dimension.}
	\label{fig:rs_col_avg}
\end{figure}

\section{Dimensionality Estimation Algorithm}\label{sec:dd_algo}
The superficial dimension of data is usually much higher than the intrinsic dimension.  The noise in real data prevent the singular values from being exactly 0.  To prevent smaller singular values or proxies of singular values from inflating the intrinsic dimension of data, we use 2 different analytic: 
\begin{enumerate}
  \item Greater Than Equal To 1\%: Count the singular values or singular values proxies that are larger than 1\% to estimate intrinsic dimension.  The threshold 1\% is configurable.(Algorithm: \ref{algo:gte_1pct})
  \item Up to 90\%: Using the largest singular values or singular values equivalent, count the number of values required such that the cumulative value is larger than 90\% of the sum of the values.  Again, 90\% threshold is configurable.(Algorithm: \ref{algo:upto_90pct})
\end{enumerate}

\begin{algorithm}[!htp]
\caption{ \textit{Dimensionality using Greater Than Equal To 1\%.
\\ Inputs: $({\sigma_1}, {\sigma_2}, ...{\sigma_n})$ - singular values or singular value proxies in case of autoencoder, threshold ($t = 1\%$).
\\Output: $p$, the number of values greater than equal 1\% . }}
\begin{algorithmic}[1]
      \State Calculate $\bs{{\sigma}_{sum}}=\bs{\sum_{i=1}^n} {\sigma_i}$
      \State Dimensionality $p = {\sum I({\sigma}_{i\%}})$, where 
        \begin{equation}\label{eqn:adj1}
          I({\sigma}_{i\%})= \begin{cases}
                        	1 	& : \text{if}~\frac{{\sigma_i}}{{\sigma}_{sum}} \geq 1\% \\
       	                      0 	& : \text{otherwise,}
          \end{cases}
        \end{equation}
\newline
\end{algorithmic}\label{algo:gte_1pct}
\end{algorithm}

\begin{algorithm}[!htp]
\caption{ \textit{Dimensionality using up to 90\%.
\\ Inputs: $({\sigma_1}, {\sigma_2}, ...{\sigma_n})$ - singular values or singular value proxies in case of autoencoder, threshold ($t = 90\%$).
\\Output: $p$, the number of largest singular values that explains 90\% of variance in ($\hat{\bs{X}}$). }}.
\begin{algorithmic}[1]
      \State Sort $({\sigma_1}, {\sigma_2}, ...{\sigma_n})$ in descending order, where $\bs{\sigma}_i$ are singular values.
      \State Calculate $\bs{{\sigma}_{sum}}=\bs{\sum_{i=1}^n} {\sigma_i}^2$
      \State Calculate ${\sigma}_{i\%}$, where ${\sigma}_{i\%} =  \frac{{\sigma_i}^2}{{\sigma}_{sum}}$
      \State Dimensionality $p$, is the value of $l$ where $\bs{\sum_{l=1}^w} {\sigma_{i\%}} \geq t(90\%)$
\newline
\end{algorithmic}\label{algo:upto_90pct}
\end{algorithm}

\section{Experiments}\label{sec:experiment}
Our experiments comprise of 2 part.  In the first part we estimate dimension of each digit in MNIST dataset.  The second part of the experiment estimates dimension of each day in US market using S\&P 500 index constituents.  While MNISTdata is static in nature, S\&P 500 dataset has varying correlation among S\&P 500 index constituents.

\subsection{Data}\label{sec:data}
\begin{enumerate}
  \item MNIST (figure~\ref{fig:expt_input}), the static data set, is a publicly available dataset comprising digits 0 to 9.  There are 60,000 training images and 10,000 testing images.  Each image is 28 by 28, which gives us 784 features.  Each feature is the pixel intensity, which is normailzed before dimension is estimated.  Since we estimate dimension of each digit for convenience, we create a derived dataset, one for each digit.
  \item S\&P 500\footnote{The S\&P 500 or Standard \& Poor's 500 Index is a market-capitalization-weighted index of the 500 largest U.S. publicly traded companies. The index is widely regarded as the best gauge of large-cap U.S. equities.} dataset (figure~\ref{fig:expt_input}) was created by downloading end of day dividend adjusted prices of S\&P 500 index constituents.  We use daily log returns over last 10 years -- January 1 2008 to December 31 2018.  When daily returns are missing we use 0 as a convenience.
\end{enumerate}

\begin{figure}[htp]
        	\centering
        	\includegraphics[width=3.2in]{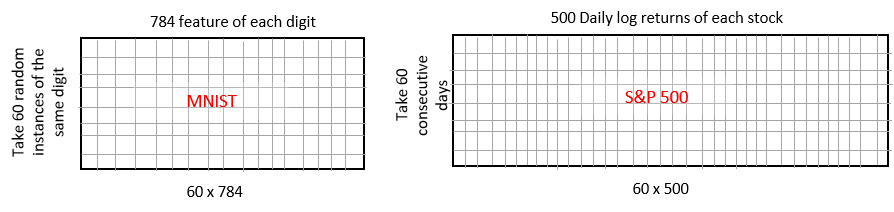}        
        	\caption{MNIST input for each iteration is randomly selected 60 instances of the digit.  Each digit has 784 features.  Russell 3000 input has 3000 log returns as its feature.  To estimate dimension at $t$, we use previous $60$ daily log returns of the stocks.}
	\label{fig:expt_input}
\end{figure}

\subsection{Estimate MNIST digit dimension}\label{sec:sge}
We calculate and analyze dimension of each digit using \emph{Greater Than Equal To 1\%} (algorithm~\ref{algo:gte_1pct}) and \emph{Up to 90\%} (algorithm~\ref{algo:upto_90pct} ).  First we use largest singular values proxies till we get 90\% of the cumulative singular values proxies to estimate dimension.  Second, we also estimate dimension using singular values proxies that are greater than 1\% of the cumulative singular values.  



\begin{table*}[ht]
    \begin{center}
        \begin{tabular}{ccccccc}
            \hline
            &\multicolumn{2}{c}{PCA} 
            &    
            \multicolumn{2}{c}{Isomap} 
            &    
            \multicolumn{2}{c}{Autoencoder} \\\hline
	MNIST Digit & Upto 90\% & $>=1\%$  & Upto 90\% & $>=1\%$  & Upto 90\% & $>=1\%$ \\\hline
      0 & 42 & 30 & 22 & 27 & 32 & 22 \\\hline
      1 & 37 & 26 & 21 & 25 & 24 & 13 \\\hline
      2 & 44 & 35 & 22 & 28 & 32 & 22 \\\hline
      3 & 43 & 34 & 23 & 28 & 31 & 19 \\\hline
      4 & 43 & 33 & 22 & 27 & 31 & 20 \\\hline
      5 & 44 & 34 & 22 & 27 & 32 & 22 \\\hline
      6 & 42 & 31 & 22 & 27 & 32 & 21 \\\hline
      7 & 42 & 31 & 22 & 27 & 30 & 19 \\\hline
      8 & 43 & 34 & 22 & 28 & 31 & 20 \\\hline
      9 & 42 & 31 & 22 & 27 & 30 & 18 \\\hline
        \end{tabular}      
    \end{center}
    \caption{Dimension of MNIST digits using PCA, Isomap and autoencoder.}
    \label{fig:all_mnist_digit_dim}
\end{table*}



\subsection{Estimate S\&P 500 dimension}\label{sec:sge}
To estimate S\&P 500 dimension we consider \emph{60 days} sliding window of daily log return of 500 stocks that comprise S\&P 500 index.  We move the window by 1 day (figure~\ref{fig:spy500_expt}), as it happens in financial markets, and again estimate dimension for next day.  We do this for all days in our dataset to get a time series of dimension.


\begin{figure*}[htp]
    \begin{subfigure}{0.5\textwidth} 
        \includegraphics[width=0.9\linewidth, height=5cm]{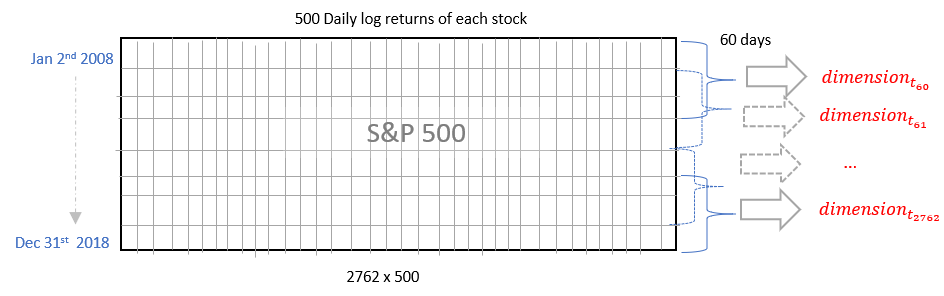} 
        \caption{Dimension time series is estimated by using 60 days of daily log returns and then moving the window by 1 day.  This gives us time series of dimension.}
        \label{fig:spy500_expt}
    \end{subfigure}
    \begin{subfigure}{0.5\textwidth} 
        \includegraphics[width=0.9\linewidth, height=5cm]{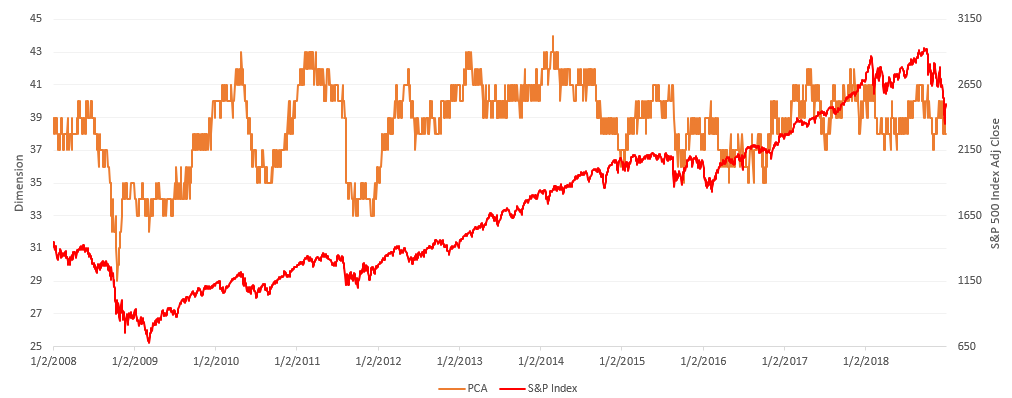}
        \caption{S\&P dimension time series using \emph{greater than 1\%} when PCA, a linear technique, is used.  The dimension of the market drops when there is stress in financial markets.}
       \label{fig:spy_dd_pca}
    \end{subfigure}
    \begin{subfigure}{0.5\textwidth} 
        \includegraphics[width=0.9\linewidth, height=5cm]{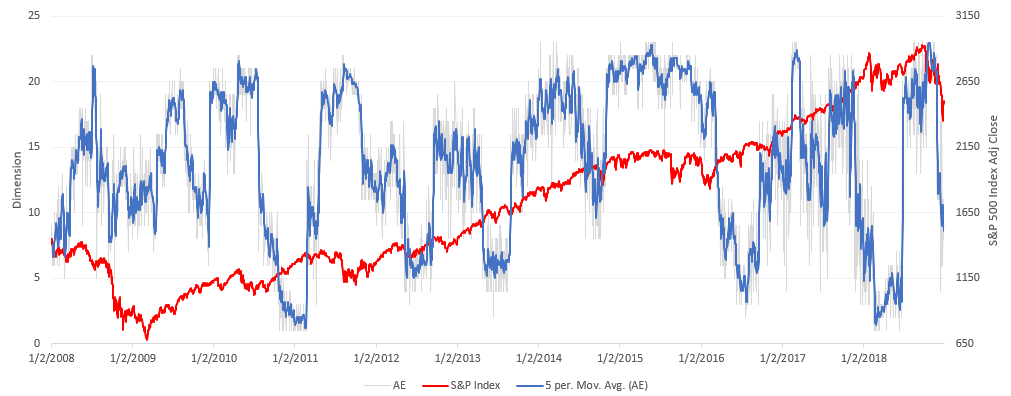}
        \caption{S\&P dimension time series can also be estimated by using autoencoder.  AE, a nonlinear technique, is highly sensitive to tuning parameters.}
       \label{fig:spy_dd_ae}
    \end{subfigure}
    \begin{subfigure}{0.5\textwidth} 
        \includegraphics[width=0.9\linewidth, height=5cm]{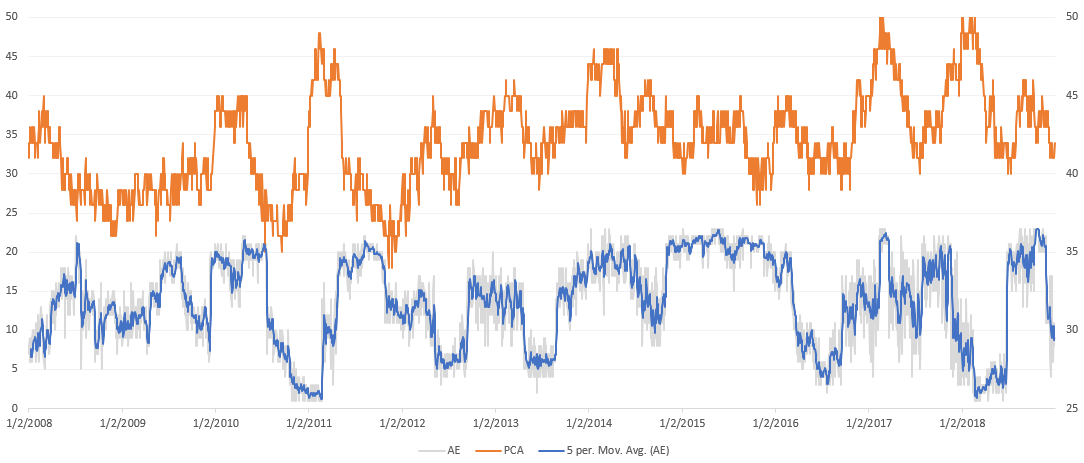}
        \caption{Both autoencoder and PCA estimate changes in dimension when there is stress in financial market.  The change in dimension is not of the same magnitude.}
       \label{fig:spy_dd_pca_vs_ae}
    \end{subfigure}
    \caption{S\&P 500 dimension estimated using PCA and autoencoder.}
\end{figure*}

\subsection{Tuning autoencoder}\label{sec:tuning_ae}
AEs have several tuning parameters.  Each parameter affects the values in hidden layer, which is used to estimate dimension. Further tuning data hungry\cite{marcus2018deep} neural network is computationally intensive. We analyzed the sensitivity of dimension estimation algorithm by iteratively experimenting with range of lambda.  Our goal was not find the optimal $\bs{\lambda}$, but to study the effect of $\bs{\lambda}$ on dimension of the dataset.  Because financial markets are stochastic, dimension of markets not only changed due to $\bs{\lambda}$ but also was amplified due to stress in market conditions.  Overtuning neural network, especially in finance, can be detrimental.  Having an hypothesis or a even broad range of financial market dimension\cite{alexeev2012equity} mitigates the tricky exercise of training neural networks with limited real financial data. 

\begin{figure*}[htp]
    \begin{subfigure}{0.5\textwidth} 
        \includegraphics[width=0.9\linewidth, height=5cm]{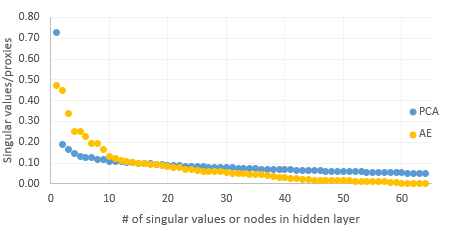} 
        \caption{We compare the singular values from PCA and hidden layer values from autoencoder for August 8th 2011 a day in market when S\&P 500 index dropped more than 6\%.}
        \label{fig:bear_08112011}
    \end{subfigure}
    \begin{subfigure}{0.5\textwidth} 
        \includegraphics[width=0.9\linewidth, height=5cm]{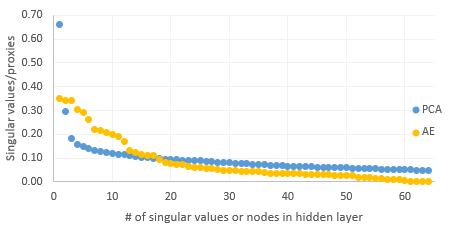}
        \caption{During October 2015 S\&P 500 index increased by 8.3\%, a large positive increase. We compare the singular values from PCA and hidden layer values from autoencoder for October 5th 2015, the day that had largest index change of 1.83\%.}
       \label{fig:bull_10052015}
    \end{subfigure}
    \caption{How does S\&P 500 dimension estimate change when there is a large up or down movement in S\&P 500, a proxy for market.}
\end{figure*}


\begin{figure*}[htp]
    \centering
    \includegraphics[width=6.4in]{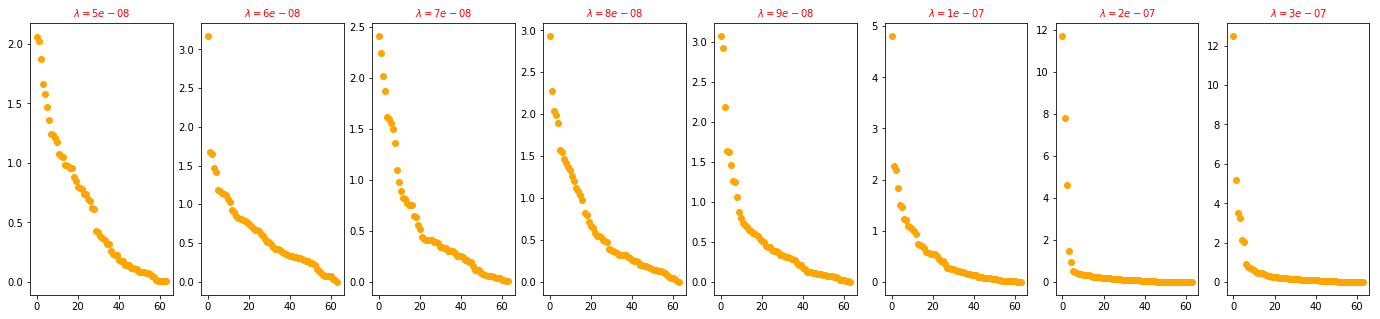}        
    \caption{We zoom into hidden layer values of August 8th 2011 and study the efffect of increasing $\lambda$ on raw hidden layer values, which is normalized to estimate dimension.}
	\label{fig:bear_08112011_raw_zlayer_diff_lambda}
\end{figure*}

\begin{figure*}[htp]
    \centering
    \includegraphics[width=6.4in]{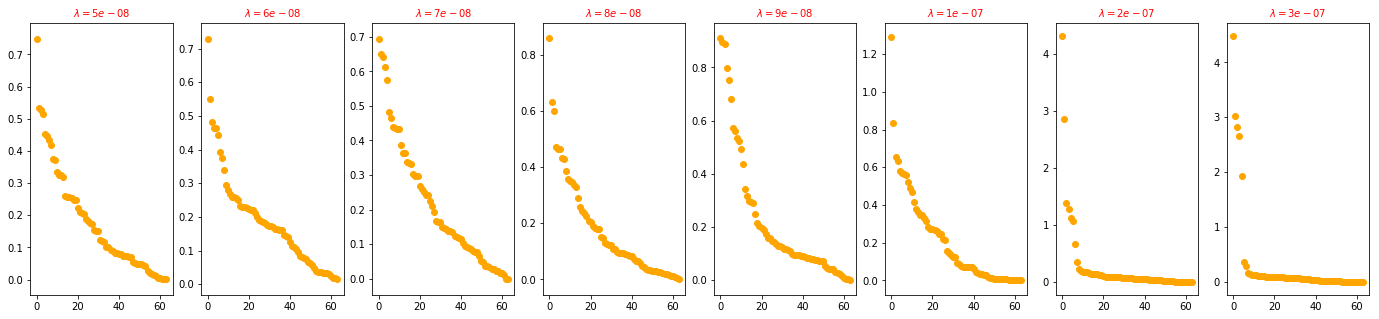}        
    \caption{S\&P 500 index climbed higher on October 5th 2015.  We zoom into the raw hidden layer values for this day and analyze the  efffect of increasing $\lambda$ on raw hidden layer values.}
	\label{fig:bull_10052015_raw_zlayer_diff_lambda}
\end{figure*}

\section{Conclusion}\label{sec:conclusion} 
This paper proposes an algorithm for estimate dimension and then uses the algorithm to estimate dimension of digits in MNIST dataset.  We find that if linear dimension reduction technique is used then dimension estimated for digits, except digit 1, is between 30 and 32.  If autoencoder is used to estimate dimension then dimension estimated for digits, except digit 1, is between 19 and 22.  S\&P 500 dimension time series from 2008 to 2018 showed higher variance.  Consistent with MNIST digits, nonlinear autoencoder dimension was lower.  Moreover, dimension estimated using autoencoder showed much higher variance compared to linear dimension estimation technique.  Our conjecture is lack of data to train data hungry autoencoder.  Further, the paper provides a set of design criteria that can be followed for transforming an autoencoder latent state representation to singular value proxies that can be used for estimating dimension.

\section*{Acknowledgement}
Results in this paper were obtained in part using a high-performance computing system acquired through NSF MRI grant DMS-1337943 to WPI.

\appendices




\end{document}